\def\year{2021}\relax
\documentclass[letterpaper]{article} %

\usepackage[title]{appendix}
\usepackage{diagbox}
\usepackage{hyperref}
\usepackage{xr,xfrac,xspace,mathtools,amsmath,amsfonts,amssymb,bm}
\usepackage{amsmath}
\usepackage{amsthm}
\usepackage[UKenglish]{babel}
\usepackage{thmtools, thm-restate}
\usepackage{accents}
\usepackage{amssymb}
\usepackage{xspace}
\usepackage{epstopdf}
\usepackage{booktabs}
\usepackage{epsfig}
\usepackage[normalem]{ulem}
\usepackage[switch]{lineno}
\usepackage[font=small]{subfig}
\usepackage[noabbrev,capitalise]{cleveref}
\usepackage[group-separator={,}]{siunitx}
\usepackage{xcolor,colortbl}
\usepackage{lipsum}
\usepackage{capt-of}
\usepackage[linesnumbered,ruled,lined,boxed]{algorithm2e}
\usepackage{algpseudocode}
\newlength{\commentWidth}
\usepackage{fancyhdr}
\usepackage{multirow}
\usepackage[normalem]{ulem}
\usepackage{soul}
\usepackage[utf8]{inputenc}
\usepackage{relsize}
\usepackage{stmaryrd}
\usepackage{wrapfig}
\usepackage{array}
\usepackage{enumitem}
\usepackage{tikz}
\usepackage{nth}
\usepackage{makecell}

\usetikzlibrary{arrows.meta,calc,backgrounds,angles,quotes}

\makeatletter

\creflabelformat{equation}{(#2#1#3)} %
\crefname{ineq}{Inequality}{Inequalities}
\creflabelformat{ineq}{(#2#1#3)}
\crefname{prob}{Problem}{Problems}
\creflabelformat{prob}{(#2#1#3)}
\crefname{seq}{Sequence}{Sequences}
\creflabelformat{seq}{(#2#1#3)}

\crefformat{section}{\S#2#1#3} %
\crefformat{subsection}{\S#2#1#3}
\crefformat{footnote}{#2\footnotemark[#1]#3}
\crefname{algocf}{Algorithm}{Algorithms}
\Crefname{algocf}{Algorithm}{Algorithms}
\crefname{theo}{Theorem}{Theorems} 
\crefname{lem}{Lemma}{Lemmas} 
\crefname{prop}{Proposition}{Propositions}
\crefname{exmpl}{Example}{Examples}
\crefname{ass}{Assumption}{Assumptions}
\crefname{rmk}{Remark}{Remarks}
\crefname{cor}{Corollary}{Corollaries}
\crefname{myalign}{Equation}{Equations}

\crefformat{app}{Appendix~#2#1#3}
\crefrangeformat{app}{Appendices~#3#1#4 to #5#2#6}
\crefmultiformat{app}{Appendices~#2#1#3}{ and #2#1#3}{, #2#1#3}{ and #2#1#3}
\crefrangemultiformat{app}{Appendices~#3#1#4 to #5#2#6}{ and #3#1#4 to #5#2#6}{, #3#1#4 to #5#2#6}{ and #3#1#4 to #5#2#6}

\newtheorem{definition}{Definition}
\newtheorem*{definition*}{Definition}
\newtheorem{theorem}{Theorem}
\newtheorem*{theorem*}{Theorem}

\newtheorem*{lemma*}{Lemma}

\newtheorem*{proposition*}{Proposition}

\newtheorem*{example*}{Example}

\newtheorem*{assumption*}{Assumption}

\newtheorem*{remark*}{Remark}

\newtheorem*{corollary*}{Corollary}

\newtheorem*{question*}{Request}

\newcommand{\nospaceitemize}{\vspace{-0.5\topsep}}

\newlist{myitemize}{itemize}{1}
\setlist[myitemize,1]{label=\textbullet,leftmargin=0.2in,wide, labelwidth=!, labelindent=0pt}

\newlist{myenumerate}{enumerate}{1}
\setlist[myenumerate,1]{label*=(\arabic*),leftmargin=0.2in,wide, labelwidth=!, labelindent=0pt}

\newenvironment{myalign*}{\par\nobreak\noindent\nonumber\align}{\endalign}
\newenvironment{myalign}{\par\nobreak\noindent\align}{\endalign}

\newcounter{letter}
\@whilenum\value{letter}<27\do{%
  \expandafter\edef\csname \Alph{letter}v\endcsname{\noexpand\bm{\Alph{letter}}}
  \expandafter\edef\csname c\Alph{letter}\endcsname{\noexpand\mathcal{\Alph{letter}}}
  \expandafter\edef\csname bb\Alph{letter}\endcsname{\noexpand\mathbb{\Alph{letter}}}
  \expandafter\edef\csname \alph{letter}v\endcsname{\noexpand\bm{\alph{letter}}}
  \stepcounter{letter}%
}

\newcommand{\alphav}{{\boldsymbol \alpha}}

\newcommand{\Dav}{{ \Delta \alphav}}

\newcommand{\thetav}{{\boldsymbol \theta}}

\newcommand{\etav}{{\boldsymbol \eta}}
\newcommand{\Exp}{\mathbb{E}}
\newcommand{\gap}{{\text{Gap}}}

\newcommand\footnoteref[1]{\protected@xdef\@thefnmark{\ref{#1}}\@footnotemark}

\allowdisplaybreaks

\newcommand{\noimage}{%
  \setlength{\fboxsep}{-\fboxrule}%
  \fbox{\phantom{\rule{10pt}{10pt}}File missing\phantom{\rule{10pt}{10pt}}}%
}
\let\includegraphicsoriginal\includegraphics
\renewcommand{\includegraphics}[2][width=\textwidth]{\IfFileExists{#2}{\includegraphicsoriginal[#1]{#2}}{\noimage}}

\newcommand*{\rom}[1]{\expandafter\@slowromancap\romannumeral #1@}

\DeclareMathOperator*{\argmin}{arg\,min}

\DeclareMathOperator{\bigO}{\mathcal{O}}

\newcommand{\rescaleEquation}[1]{\resizebox{\ifdim\width>\columnwidth
        \columnwidth
      \else
        \width
      \fi
    }{!}{$
    \displaystyle #1$}}

\usepackage{aaai21}  %
\usepackage{times}  %
\usepackage{helvet} %
\usepackage{courier}  %
\usepackage{natbib}  %
\usepackage{caption} %
\renewcommand\paragraph{\@startsection{paragraph}{4}{\z@}
                                    {1.5ex \@plus0.5ex \@minus0.5ex}
                                    {-1em}
                                    {\normalfont\normalsize\bfseries}}

\makeatother

\newcommand{\dpsgd}{\textsc{\mbox{DP-SGD}}\xspace}
\newcommand{\dpscd}{\textsc{\mbox{DP-SCD}}\xspace}
\newcommand{\dpsscd}{\textsc{\mbox{seqDP-SCD}}\xspace}
\newcommand{\pdpscd}{\textsc{\mbox{primalDP-SCD}}\xspace}
\title{Differentially Private Stochastic Coordinate Descent}

\newcounter{alley}
\newcommand{\footremember}[1]{%
    \footnote{#1}
    \setcounter{alley}{\value{footnote}}%
}
\newcommand{\footrecall}[1]{%
    \footnotemark[\value{#1}]%
} 
\author{
    Georgios Damaskinos,\textsuperscript{\rm 1}\footremember{Work partially conducted while at IBM Research, Zurich.} Celestine Mendler-D{\"u}nner,\textsuperscript{\rm 2}\footrecall{alley} \\Rachid Guerraoui,\textsuperscript{\rm 1} Nikolaos Papandreou,\textsuperscript{\rm 3} Thomas Parnell\textsuperscript{\rm 3}
    \\
}

\affiliations{
    \textsuperscript{\rm 1}École Polytechnique Fédérale de Lausanne (EPFL), Switzerland\\
    \textsuperscript{\rm 2}University of California, Berkeley\\
    \textsuperscript{\rm 3}IBM Research, Zurich\\
    georgios.damaskinos@epfl.ch, mendler@berkeley.edu, rachid.guerraoui@epfl.ch, \{npo,tpa\}@zurich.ibm.com\\
}

\begin{document}

\maketitle

\begin{abstract}

In this paper we tackle the challenge of making the stochastic coordinate descent algorithm differentially private. 
Compared to the classical gradient descent algorithm where updates operate on a single model vector and controlled noise addition to this vector suffices to hide critical information about individuals, stochastic coordinate descent crucially relies on keeping auxiliary information in memory during training. 
This auxiliary information provides an additional privacy leak and poses the major challenge addressed in this work. 
Driven by the insight that under independent noise addition, the consistency of the auxiliary information holds in expectation, we present \dpscd, the first differentially private stochastic coordinate descent algorithm. We analyze our new method theoretically and argue that decoupling and parallelizing coordinate updates is essential for its utility. On the empirical side we demonstrate competitive performance against the popular stochastic gradient descent alternative (DP-SGD) while requiring significantly less tuning. 
\end{abstract}

\section{Introduction}

Stochastic coordinate descent (SCD)~\cite{wright2015coordinate} is an appealing  optimization algorithm. 
Compared with the classical stochastic gradient descent (SGD), SCD does not require 
a learning rate to be tuned and can often have favorable convergence behavior~\cite{fan2008liblinear,dunner2018snap,ma2015adding,hsieh2015passcode}. 
In particular, for training generalized linear models, SCD is the algorithm of choice for many applications and has been implemented as a default solver in several popular packages such as Scikit-learn, TensorFlow and Liblinear~\cite{fan2008liblinear}.

However, SCD is not designed with privacy concerns in mind: 
 SCD builds models that may leak sensitive information regarding the training data records.
This is a major issue for privacy-critical domains such as health care where models are trained based on the medical records of patients.   

Nevertheless, the low tuning cost of SCD is a particularly appealing property for privacy-preserving machine learning (ML).
Hyperparameter tuning costs not only in terms of additional computation, but also in terms of spending the privacy budget.

Our goal is to extend SCD with mechanisms that preserve data privacy, and thus enable the reuse of large engineering efforts invested in SCD~\cite{parallelCD,syscd}, for privacy-critical use-cases.
In this paper we therefore ask the question: \emph{Can SCD maintain its benefits (convergence guarantees, low tuning cost) alongside strong privacy guarantees?}

We employ differential privacy (DP)~\cite{dwork2014algorithmic} as our mathematical definition of privacy.
DP provides a formal guarantee for the amount of leaked information regarding participants in a database, given a strong adversary and the output of a mechanism that processes this database.
DP provides a fine-grained way of measuring privacy across multiple subsequent executions of such mechanisms and is thus well aligned with the iterative nature of ML algorithms.

The main challenge of making SCD differentially private is that an efficient implementation stores and updates not only the model vector $\alphav$ but also an auxiliary vector $\vv:=\Xv\alphav$ to avoid recurring computations. 
These two vectors are coupled by the data matrix $\Xv$ and need to be  \emph{consistent} for standard convergence results to hold. 
However, to provide rigorous privacy guarantees, it is vital to add independent noise to both vectors which prohibits this consistency.

\paragraph{Contributions.} 
We present \dpscd, a differentially private version of the SCD algorithm~\cite{shalev2013stochastic} with formal privacy guarantees. 
In particular, we make the following contributions.
\begin{itemize}[noitemsep,topsep=0pt,leftmargin=0.5cm]
\item We extend SCD to compute each model update as an aggregate of independent updates computed on a random data sample. This parallelism is crucial for the utility of the algorithm because it reduces the amount of the noise per sample that is necessary to guarantee DP.

\item We provide the first analysis of SCD in the DP setting and derive a bound on the maximum level of noise that can be tolerated to guarantee a given level of utility. 

\item We empirically show that for problems where SCD has closed form updates, \dpscd achieves a better privacy-utility trade-off compared to the popular \dpsgd algorithm~\cite{abadi2016deep} while, at the same time, being free of a learning rate hyperparameter that needs tuning. 
Our implementation is available\footnote{\url{https://github.com/gdamaskinos/dpscd}}.

\end{itemize}

\section{Preliminaries} \label{sec:prelim}

Before we dive into the details of making SCD differentially private, we first formally define the problem class considered in this paper and provide the necessary background on SCD and differential privacy.

\subsection{Setup} \label{sec:setup}
We target the training of Generalized Linear Models (GLMs), the class of models to which SCD is most commonly applied.
This class includes convex problems of the following form:
\begin{equation} \label[prob]{eq:obj}
\small
\min_{\thetav} \cF(\thetav;\Xv) := \min_{\thetav} \left\{ \frac 1 N \sum_{i=1}^N \ell_i( \xv_i^\top \thetav) + \frac \lambda 2 \|\thetav\|^2 \right\}
\end{equation}
The model vector $\thetav\in\bbR^M$ is learnt from the training dataset ${\Xv \in \bbR^{M \times N}}$  that contains the $N$ training examples ${\xv_i\in\bbR^M}$ as columns, $\lambda>0$ denotes the regularization parameter, and $\ell_i$ the convex loss functions. 
The norm $\|\cdot\|$ refers to the $L_2$-norm. 
For the rest of the paper we use the common assumption that for all $i\in[N]$ the data examples $\xv_i$ are normalized, i.e., $\|\xv_i\| = 1$, and that the loss functions $\ell_i$ are $1/\mu$-smooth.
A wide range of ML models fall into this setup including ridge regression and $L_2$-regularized logistic regression~\cite{shalev2013stochastic}.

\paragraph{Threat model.}
We consider the classical threat model used in~\cite{abadi2016deep}.
In particular, we assume that an adversary has white-box access to the training procedure (algorithm, hyperparameters, and intermediate output) and can have access even to $\Xv\backslash \xv_k$, where $\xv_k$ is the data instance the adversary is targeting.
However, the adversary cannot have access to the intermediate results of any update computation. 
We make this assumption more explicit in \cref{sec:privacy-analysis}.

\subsection{Primal-Dual Stochastic Coordinate Descent} \label{sec:scd}
The primal SCD algorithm repeatedly selects a coordinate $j\in[M]$ at random, solves a one dimensional auxiliary problem and updates the parameter vector $\thetav$:
\begin{equation} \label[prob]{eq:scd}
\small
\thetav^+ \leftarrow \thetav + \ev_j \zeta^\star \quad \text{where} \; \zeta^\star = \argmin_{\zeta} \cF(\thetav+\ev_j\zeta;\Xv) 
\end{equation}
where $\ev_j$ denotes the unit vector with value 1 at position $j$.
\cref{eq:scd} often has a closed form solution; otherwise $\cF$ can be replaced by its second-order Taylor approximation.

A crucial approach for improving the computational complexity of each SCD update is to keep an auxiliary vector $\vv:=\Xv^\top \thetav$ in memory. This avoids recurring computations, as the loss function accesses the data $\Xv$ through the linear map $\Xv^\top \thetav$ (see \cref{eq:obj}). 
The auxiliary vector is then updated in each iteration as $\vv^+ \leftarrow \vv+\zeta^\star \xv_j$. 

\paragraph{Dual SCD.}
SCD can be equivalently applied to the dual formulation of \cref{eq:obj}, commonly referred to as SDCA~\cite{shalev2013stochastic}. 
The dual optimization problem has the following form: 
\begin{myalign} \label[prob]{eq:objD}
\small
\min\limits_{\alphav}\cF^*( \alphav; \Xv) := \min\limits_{\alphav} \left\{ \frac 1 N \sum_{i=1}^N \ell^*_i(-\alpha_i) + \frac{\|X\alphav\|^2}{2\lambda N^2}  \right\}
\end{myalign}
where $\bm{\alpha} \in \bbR^N$ denotes the dual model vector and $\ell^*_i$ the convex conjugate of the loss function $\ell_i$. 
Since the dual objective (\cref{eq:objD}) depends on the data matrix through the linear map $\Xv\alphav$, the auxiliary vector is naturally defined as $\vv:=\Xv\alphav$ in SDCA. 
We use the first order optimality conditions to relate the primal and the dual model vectors which results in  $\thetav(\alphav)= \frac 1 {\lambda N} \Xv\alphav$ and leads to the important definition of the duality gap~\cite{primal-dual}:
\begin{myalign} \label{eq:gap}
\small
\gap(\alphav) &:= \cF^*(\alphav;\Xv) + \cF(\thetav(\alphav);\Xv) \nonumber \\
&= \langle \Xv \alphav ,\thetav(\alphav) \rangle + \frac \lambda {2} \|\thetav(\alphav)\|^2 +  \frac \lambda 2 \|\thetav\|^2
\end{myalign}

By the construction of the two problems, the optimal values for the objectives match in the convex setting and the duality gap attains zero~\cite{shalev2013stochastic}. 
Therefore, the model $\thetav$ can be learnt from solving either \cref{eq:obj} or \cref{eq:objD}, where we use the map $\thetav(\alphav)$ to obtain the final  solution when solving \cref{eq:objD}.

While the two problems have similar structure, they are quite different from a privacy perspective due to their data access patterns. 
When applied to the dual, SCD computes each update by processing a single example at a time, whereas the primal SCD processes one coordinate across all the examples. 
Several implications arise as differential privacy is defined on a per-example basis.

\subsection{Differentially Private Machine Learning}

Differential privacy is a guarantee for a function $f$ applied to a database of sensitive data~\cite{dwork2014algorithmic}.
In the context of supervised ML, this function is the update function of the algorithm and the data is typically a set of input-label pairs ($\xv_i, y_i$) that are used during model training. 
Two input datasets are \emph{adjacent} if they differ only in a single input-label pair. Querying the model translates into making predictions for the label of some new input.

\begin{definition}[Differential privacy] \label{df:dp}
A randomized mechanism $\cM: \cD \rightarrow \bbR$ satisfies ($\epsilon$, $\delta$)-DP if for any two adjacent inputs $d, d' \in \cD$ and for any subset of outputs $S \subseteq \bbR$ it holds that: 
$\mathrm{Pr}[\cM(d) \in S] \leq e^\epsilon\, \mathrm{Pr}[\cM(d') \in S] + \delta$.
\end{definition}

\paragraph{The Gaussian mechanism.}
The \emph{Gaussian mechanism} is a popular method for making a deterministic function ${f: \cD \rightarrow \bbR}$ differentially private. By adding Gaussian noise to the output of the function we can hide particularities of individual input values. 
The resulting mechanism is defined as: ${\cM(d) := f(d) + \cN(0, S_f^2 \sigma^2)}$ where the variance of the noise needs to be chosen proportional to the sensitivity $S_f$ of the function $f$.
This definition can be readily extended to the multi-dimensional case in order to fit the general ML setting: An iterative ML algorithm can be viewed as a function $\fv: \bbR^{M \times N} \rightarrow \bbR^M$ that repeatedly computes model updates from the data and thus requires the addition of multi-dimensional noise at each iteration $t$: 
\begin{equation}
\small
{\bm{\cM}_t(\dv) = \fv(\dv) + \cN(0, S_f^2 \sigma^2\Iv), \; \Iv \in \bbR^{M \times M}}
\label{eq:gaussianmech}
\end{equation} 
where $\Iv$ denotes the identity matrix and the sensitivity is:
\begin{equation} \label{eq:sensitivity}
\small
S_f := \max\limits_{\text{adjacent} \;\dv,\dv'} \|\fv(\dv) - \fv(\dv')\|
\end{equation}

\paragraph{\dpsgd.}
As an application of the Gaussian mechanism, we consider stochastic gradient descent (SGD) used for solving \cref{eq:obj}.
SGD iteratively updates the model as ${\thetav^+ \leftarrow \thetav - \eta \bm{g}_\xi}$ with ${\bm{g}_\xi:= \frac{1}{|\xi|}\sum_{i \in \xi}\nabla \ell_i (\xv_i^\top \thetav) \xv_i + \lambda  \thetav}$ being the gradient computed on a dataset sample $\xi$ (also know as \emph{mini-batch}), and $\eta$ denoting the learning rate. 
According to the Gaussian mechanism, \dpsgd adds noise to each update $\bm{g}_\xi$ with the variance given by the sensitivity of the update function, i.e., ${S_f=\|\gv_\xi\|}$ for our example. 
In practice, an additional gradient clipping step enforces a desired bound on the sensitivity $S_f$~\cite{abadi2016deep}.

\paragraph{Privacy accounting.} 
\label{sec:accountant}
Measuring the privacy leakage of a randomized mechanism $\cM$ boils down to computing ($\epsilon, \delta$). More precisely, computing a bound for the privacy loss $\epsilon$ that holds with probability $1-\delta$.
In the context of ML, $\cM$ often consists of a sequence of mechanisms $\cM_t$ that, for example, denote the model update at each iteration $t$. 
All these mechanisms have the same pattern in terms of sensitive (training) data access for most iterative ML algorithms, including SCD. Computing ($\epsilon, \delta$) given the individual pairs ($\epsilon_t, \delta_t$) is a problem known as composability. 
However, standard composition theorems often provide loose bounds on ($\epsilon, \delta$) and methods such as the \emph{moments accountant}~\cite{abadi2016deep} are useful tools to compute tighter bounds for the overall privacy loss.
The moments accountant method is tailored to the Gaussian mechanism and employs the log moment of each $\cM_t$ to derive the bound of the total privacy loss.
It can be viewed as a function that returns the privacy loss bound: 
\begin{equation} \label{eq:MA}
\epsilon = \text{MA}(\delta, \sigma, q, T)
\end{equation}
where $\sigma$ is the noise magnitude, $q$ is the sampling ratio (i.e., the ratio of the data that each $\cM_t$ uses over the total data), and $T$ is the number of individual mechanisms $\cM_t$.

\begin{algorithm}[t]
\caption{\dpscd (for \cref{eq:objD})} \label{algo:dp-pscd_dual}
\small
\DontPrintSemicolon
\KwIn{$N$ examples $\xv_i\in\bbR^M$ and the corresponding labels $y_i$, $T$: \# iterations,  $L$: mini-batch size, $(\epsilon, \delta)$: DP parameters, $C$: scaling factor}
\textbf{Init:} $\alphav = \bm{0}$ ; $\vv = \bm{0}$ ; shuffle examples $\xv_i$ \;
$\sigma \leftarrow \textit{smallest noise magnitude, s.t., } \text{MA}\left(\delta, \sigma, \frac L N , T\right) = \epsilon$  

\For{$t = 1,2,...T $}{
  $\Delta \alphav=0$ ; $\Delta \vv=0$\;
  Sample a mini-batch $\mathcal{B} \subset [N]$ of $L$ examples\;
  \For{$j \in \cB$}{

    $\zeta_j = \argmin_\zeta \cG_j^*(\alpha_j,\zeta, \vv, \xv_j)$ \; 
    $\zeta_j \mathbin{{/}{=}} \max\left(1, \frac{|\zeta_j|}{C}\right)$ \tcp*[r]{scale}

    $\Delta\alphav\; \mathbin{{+}{=}} \ev_j \zeta_j$ \;

    $\Delta \vv \; \mathbin{{+}{=}} \zeta_j \xv_j $ \;

   }

  \tcp*[l]{update the model} 
  $\alphav\; \mathbin{{+}{=}} \Delta\alphav + \ev_\cB (\cN(0, \sigma^2 2 C^2 \bm{I_L}))$, \quad $\bm{I_L} \in \bbR^{L \times L}$ \;
  $\vv\; \mathbin{{+}{=}} \Delta\vv + \cN(0, \sigma^2 2 C^2 \bm{I_M})$, \quad $\bm{I_M} \in \bbR^{M \times M}$ \;

}
\Return $\bm{\theta} =\frac 1 {\lambda N}\vv$ \tcp*[r]{primal DP model}
\end{algorithm}

\section{Differentially Private Stochastic Coordinate Descent} \label{sec:dpscd}

We first focus on \cref{eq:objD} and discuss the primal problem in \cref{sec:primal}. 
\cref{algo:dp-pscd_dual} summarizes our differentially private SCD algorithm (\dpscd). 
The crucial extension in comparison with standard SDCA is that we consider updates that independently process a mini-batch $\cB$ of $L$ coordinates sampled uniformly at random in each inner iteration (Steps~6-11). 
This is not only beneficial from a performance perspective, as the updates can be executed in parallel, but it also serves as a hyperparameter that can improve the privacy-utility trade-off of our algorithm (similar to the lot size in~\cite{abadi2016deep}).
The subproblem formulation $\cG_j^*$ in Step~7 is inspired by the distributed coordinate descent method of~\citet{ma2015adding}. 
In particular, we reuse the local subproblem formulation of their method for the case where each parallel step updates only a single ${j\in\cB}$:
\begin{myalign}\label{eq:Fhat}
\small
\cG_j^*(\alpha_j,&\zeta, \vv, \xv_j):=\frac 1 N \ell^*_j(-\alpha_j-\zeta) \nonumber \\
&+ \frac 1 {2\lambda N^2} \left(  \|\vv\|^2 + 2 \xv_j^\top \vv \zeta+  L  \|\xv_j \|^2 \zeta^2 \right)
\end{myalign}

In each iteration $t>0$ we then minimize \cref{eq:Fhat} over $\zeta$  for all  ${j\in\cB}$ to obtain the respective coordinate updates $\zeta_j$.
This minimization can often be computed in closed form, e.g., for ridge regression, or SVMs. 
However, approximate solutions are sufficient for convergence, e.g., as specified in \cite[Assumption~1]{smith2017cocoa}.
For logistic regression we use a single Newton step to approximate the coordiante update $\zeta_j$.

Finally, to bound the sensitivity of each update step we rescale $\zeta_j$ to have magnitude no larger than $C$ (Step~8). We can then use the Gaussian mechanism to make $\alphav$ and $\vv$ differentially private (Steps~12-13).
We address two main questions regarding \dpscd: 
\begin{enumerate}[noitemsep,topsep=5pt,leftmargin=*]
\item How much noise do we need to guarantee $(\epsilon,\delta)$-DP?
\item Can we still give convergence guarantees for this new algorithm under the addition of noise?
\end{enumerate}
We answer the first question in \cref{sec:privacy-analysis} by analyzing the sensitivity of our update function with respect to $(\alphav,\vv)$.
For the second question, the main challenge is that after the addition of noise the consistency between $\alphav$ and $\vv$, i.e., $\vv = \Xv\alphav$, no longer holds. In \cref{sec:analysis} we  show how to address this challenge and prove convergence for our method.
We provide the cost analysis and implementation details related to \dpscd in \cref{sec:appendix_cost} and \cref{sec:appendix_implementation} respectively.

\subsection{Privacy Analysis}
\label{sec:privacy-analysis}

We view the training procedure of \cref{algo:dp-pscd_dual} as a sequence of mechanisms $\bm{\cM}_t$ where each mechanism corresponds to one outer iteration (Steps~3-14) and computes an update on a mini-batch of  $L$ examples. 
We assume these mechanisms to be atomic from an adversary point of view, i.e., we assume no access to the individual coordinate updates within the mini-batch $\cB$. 
The update computation (Steps~6-11) before the noise addition, corresponds to $\fv$ according to \cref{eq:gaussianmech}.
For determining the sensitivity of $\fv$ it is important to note that all updates within a mini-batch $\cB$ touch different data points and are computed independently. 
The output of each mechanism $\bm{\cM}_t$ is the concatenation $[\alphav; \vv]$ of the updated model vector $\alphav$  and the updated auxiliary vector $\vv$. 
The sensitivity of this mechanism is given as follows.

\begin{restatable}[Sensitivity of \dpscd]{lem}{sensdppscd} \label{lemma:sens-dppscd} 
Assume the columns of the data matrix $\Xv$ are normalized.
Then, the sensitivity of each mini-batch update computation (Steps~6-11 in \cref{algo:dp-pscd_dual}) is bounded by: ${S_f \leq \sqrt{2}C}$.
\end{restatable}

\begin{proof}[Proof Sketch]
The sensitivity depends on $\fv(\Xv), \fv(\Xv')$ that differ solely due to the missing example for $\Xv'$, as the updates for a given mini-batch are independent.
The bound follows from Step~8, the triangle inequality and the assumption of normalized data.
The full proof can be found in \cref{sec:appendix_dppscd_sens}.
\end{proof}

\begin{theorem}[Privacy bound for \dpscd] \label{thm:sigmaBound}
\cref{algo:dp-pscd_dual} is \emph{($\epsilon, \delta$)-differentially private} for any $\epsilon = \bigO\left(q^2 T\right)$ and $\delta > 0$ if we choose $\sigma = \Omega\left(\frac 1 \epsilon  q \sqrt{T \, \log(1/\delta)}\right)$.
\end{theorem}

\begin{proof}
Each mechanism $\bm{\cM}_t$ is made differentially private by using the Gaussian mechanism from \cref{eq:gaussianmech}. The output dimension is $M+L$ in our case.  
The moments of each mechanism $\bm{\cM}_t$ are bounded given \cref{lemma:sens-dppscd} and~\cite[Lemma~3]{abadi2016deep}.
Hence, based on~\cite[Theorem~1]{abadi2016deep}, we derive the proposed lower bound on $\sigma$ that guarantees ($\epsilon, \delta$)-DP for the output model.
\end{proof}

\subsection{Primal Version} \label{sec:primal}

The primal formulation of \dpscd (\cref{algo:dp-pscd_primal} in \cref{sec:appendix_primal}) computes the updates in a coordinate-wise manner, thus making differentially private learning more challenging than in the dual formulation.
For each update computation, the primal version accesses a single coordinate $j$ across all examples. 
This means the sampling ratio $q$ for each mechanism $\cM_t$ is equal to 1 and the number of parallel updates ($L$) is no longer an effective parameter to regulate the privacy-utility trade-off.
Additionally, using the Gaussian mechanism, the sensitivity bound, and thus the noise addition necessary for DP, is significantly larger (as $S_f \leq 2C \sqrt{L(L+1)}$) compared to the dual version (see \cref{lemma:sens-dppscd_primal} in \cref{sec:appendix_primal}).
In conclusion, the dual version (\cref{algo:dp-pscd_dual}) is preferable over the primal.

\section{Convergence Analysis}
\label{sec:analysis}

We recall that the main challenge for generalizing the convergence guarantees of SCD to \dpscd, is how to handle potential inconsistencies between the auxiliary vector $\vv$  and the model vector $\alphav$.
Note that a variant of \cref{algo:dp-pscd_dual} that only updates $\alphav$ and recomputes $\vv$ in every iteration would overcome this issue. 
However, such a variant involves several disadvantages that make it highly impractical.
First it involves a significant computational overhead.
Second it makes the mini-batch size ineffective as a tuning parameter, as the sampling ratio ($q$ in \cref{eq:MA}) would be 1 for each update.
Third, on the final step it would need to employ the entire dataset to map the dual model to the primal model ($\thetav:=\frac 1 {\lambda N}\bm{X \alphav}$), which creates a massive privacy leakage. 
All these can be avoided by maintaining the auxiliary vector and computing the final primal model as $\thetav:=\frac 1 {\lambda N}\vv$. %

To analyze the convergence of \cref{algo:dp-pscd_dual} we split each mini-batch iteration in two steps: an update step and a perturbation step. 
We denote the privacy preserving model sequence by $\{\alphav_t\}_{t>0}$ and the intermediate, non-public models before the perturbation step by $\{\hat\alphav_t\}_{t>0}$.
We use the same notation for the corresponding $\vv$ sequences.
The \emph{update step} includes the computation of $L$ coordinate updates (i.e., $\hat \alphav_{t-1}= \alphav_{t-1} + \Delta \alphav_t$ and $\hat\vv_{t-1} = \vv_{t-1} + \Delta \vv_t$) and the \emph{perturbation step} adds Gaussian noise to the two vectors $\alphav$ and $\vv$ independently  (i.e., $\alphav_t= \hat \alphav_{t-1} + \etav_\alpha$ and $\vv_t = \hat \vv_{t-1} + \etav_v$, where $\etav$ denotes the Gaussian noise). 
Hence, we get the following sequence:
\begin{equation*}
\small
\ldots \rightarrow\{\alphav_{t-1},\vv_{t-1}\}\overset{(\text{update})}{\rightarrow} \{\hat \alphav_{t-1},\hat \vv_{t-1}\}\overset{(\text{perturb})}{\rightarrow} \{\alphav_{t},\vv_{t}\}\rightarrow \ldots
\end{equation*}

Our approach is to show that the update step decreases the objective even if the update is computed based on a noisy version of $\alphav, \vv$ and the amount of decrease is larger than the damage caused by adding noise in the perturbation step.
The key observation that allows us to derive convergence guarantees in this setting is the following.
\begin{restatable}[Consistency in expectation]{rmk}{sscdconsistency} \label{remark:pscd_consistency}
Given the construction of the model updates and the independent noise with zero mean that is added to both sequences, \cref{algo:dp-pscd_dual} preserves the consistency between $\alphav$ and $\vv$ in expectation, i.e., $\Exp[\vv] = \Xv \Exp[\alphav]$.
\end{restatable}

\subsection{Update Step}
Each iteration of \cref{algo:dp-pscd_dual} computes a mini-batch update $\Dav$ that is applied to the model $\alphav$ and indirectly to the auxiliary vector $\vv$ in Steps~12 and 13, respectively. 
We denote by $\Delta\alphav^\text{tmp}$ the unscaled version of this update, i.e., the update computed excluding Step~8.
We add this step back later in our analysis.
\cref{lemma:pscd_lower} gives a lower bound for the decrease in the objective achieved by performing this update even if $\Delta\alphav^\text{tmp}$ is computed based on noisy versions of $\alphav, \vv$ where $\Exp[\vv]=\Exp[\Xv\alphav]$ but $\vv\neq \Xv\alphav$.

\begin{restatable}[Update step - objective decrease lower bound]{lem}{pscdlower} \label{lemma:pscd_lower}
Assuming $\ell_i$ are $1 / \mu$-smooth,
then the update step of \cref{algo:dp-pscd_dual} decreases the objective, even if computed based on a noisy version of $\alphav, \vv$. The decrease is lower-bounded as follows:
\begin{equation} \label[ineq]{eq:pscdlower}
\Exp[\cS(\alphav) - \cS(\alphav+  \Delta\alphav^\mathrm{tmp}) ]\geq \frac{\mu \lambda L}{\mu \lambda N+L} \Exp[\cS(\alphav)] 
\end{equation}
where $\cS$ denotes the dual suboptimality defined as: ${\cS(\alphav):= \cF^*(\alphav;\Xv)-\min_\alphav \cF^*(\alphav;\Xv)}$.
\end{restatable}

\begin{proof}[Proof Sketch] 
We build on~\cite[Lemma~3]{ma2015adding} to relate the decrease of each parallel update computed on the subproblems $\cG_j^*$ to the global function decrease. Then, we take expectation w.r.t the randomization of the noise  and   proceed along the lines of~\cite[Lemma~5]{ma2015adding} to involve  the duality gap in our analysis.
Finally, based on an inequality for the duality gap and \cref{remark:pscd_consistency} we arrive at the bound stated in \cref{lemma:pscd_lower}.
Note that with $L=1$ we recover the classical result of SDCA~\cite{shalev2013stochastic} for the sequential case. The full proof can be found in \cref{sec:proof_pscd_lower}.
\end{proof}

\paragraph{Incorporating update scaling.}
When computing the update $\Dav$ in \cref{algo:dp-pscd_dual}, each coordinate of $\Delta\alphav^\text{tmp}$ is \emph{scaled} to a maximum magnitude of $C$ (Step~8) in order to bound the sensitivity of each update step. 
In strong contrast to SGD, where this scaling step destroys the unbiasedness of the gradients and thus classical convergence guarantees no longer hold, for \dpscd the scaling only translates into a smaller function decrease.
This is a remarkable property of SCD when analyzed in the DP setting.

To incorporate scaling into our analysis we use the following inequality which is guaranteed to hold for some $\kappa\in[0,1)$ due to the convexity of the objective.
\begin{myalign} \label[ineq]{eq:pscdscaling}
\small
&\cS(\alphav+ \Delta\alphav) \leq (1-\kappa)\cS(\alphav+\Delta\alphav^\text{tmp}) +\kappa \cS(\alphav)  \nonumber \\ 
&\Leftrightarrow \Exp[\cS(\alphav) - \cS(\alphav+ \Delta\alphav)] \geq \nonumber \\
&\qquad\qquad(1-\kappa)\Exp[\cS(\alphav) - \cS(\alphav+\Delta\alphav^\text{tmp})]
\end{myalign}
The scaling step preserves the linear convergence of \cref{lemma:pscd_lower} and decreases the lower-bound on the RHS of \cref{eq:pscdlower} by a factor of $(1-\kappa)$. Note that for $\kappa=0$ (i.e., no scaling) the solution is exact, and  
the smaller the scaling factor $C$, the larger the $\kappa$.

\subsection{Perturbation Step}
To derive a utility guarantee for \dpscd, it remains to show that adding noise at the end of each mini-batch update does not increase the objective more than the decrease achieved by the rescaled update $\Delta\alphav$.

\begin{restatable}[Perturbation step - objective increase upper bound]{lem}{pscdupper} \label{lemma:pscd_upper}
The perturbation step of \cref{algo:dp-pscd_dual} increases the objective by at most:
\begin{equation} \label[ineq]{eq:pscdupper}
\small
\Exp[|\cS(\alphav+ \Delta\alphav +\etav )-\cS(\alphav + \Delta\alphav)|] \leq  \frac {L \sigma^2} {2 \lambda N^2} 
\end{equation}
\begin{proof}

Given the $L_2$ regularization, $\cF$ is $\lambda$-strongly convex for any convex $\ell_i$. 
By reusing the Fenchel-Rockafellar duality pair presented by~\citet{primal-dual} (see Equation~(A)-(B) in their work) and incorporating the $1/N$ rescaling of \cref{eq:objD}, we get that $\cF^*$ is $\frac{1}{\lambda N^2}$-smooth.
We thus have ${\cF^*(\alphav' + \etav )\leq\cF^*(\alphav')+ \etav^\top \nabla \cF^*(\alphav') + \frac {1}{2 \lambda N^2} \|\etav\|^2}$. 
Setting $\alphav'=\alphav+\Delta\alphav$, subtracting $\min_a \cF^*(\alphav)$ on both sides and taking expectations w.r.t the randomness in the perturbation noise,
the claim follows from $\Exp[\etav]=0$ and $\Exp[\|\etav\|^2]=L\sigma^2$.
\end{proof}
\end{restatable}

Finally, we combine the different steps from our analysis (\cref{lemma:pscd_lower}, \cref{lemma:pscd_upper}, and \cref{eq:pscdscaling}) with the privacy bound from \cref{thm:sigmaBound} and derive our main result stated in \cref{thm:dpscd_conv}.
The proof can be found in \cref{sec:proof_pdscd_conv}.

\begin{restatable}[Utility guarantee for \cref{algo:dp-pscd_dual}]{theo}{pscdconv} \label{thm:dpscd_conv}
Suppose that $\ell_i$ is convex and $1 / \mu$-smooth $\forall i$.
If we choose $L$, $C$ such that ${L (2(1-\kappa)\mu\lambda - 1) > \mu\lambda N}$ for ${\kappa\in(0,1)}$, and $T$ such that ${T = \bigO\left(\mathrm{log}\left(\frac{\lambda N^4 \epsilon^2}{L^3 \; \log(1/\delta)}\right)\right)}$, then the suboptimality of \cref{algo:dp-pscd_dual} is bounded as:
\begin{equation} \label[ineq]{eq:main_theo}
\rescaleEquation{
\Exp[\cS(\alphav^{(T)})]  \leq \bigO\left(\frac{L^3}{\lambda N^4\epsilon^2} \mathrm{log}\left(\frac{\lambda N\epsilon}{L}\right)  \mathrm{log}\left(\frac 1 \delta\right)\right)}
\end{equation}
where $L/N$ is the sampling ratio $q$.
\end{restatable}

The suboptimality is proportional to the magnitude of the noise and hence finding the exact minimizer requires ${\sigma\rightarrow 0}$ (i.e., ${\epsilon\rightarrow \infty}$).
The smaller the $\sigma$ the larger the $\epsilon$ and thus the less private the learning is.
We empirically confirm that \dpscd converges smoother with a smaller $\sigma$ in \cref{sec:exp}.

Noteworthy, there is no lower bound on the suboptimality.
DP-SCD converges to a ball around the optimum and fluctuates randomly within this ball.
The radius of this ball depends on $\sigma$ and constitutes our utility loss upper bound (\cref{thm:dpscd_conv}). 
The utility is equal to this radius when the function decrease (\cref{lemma:pscd_lower} after incorporating update scaling) is equal to the increase by the perturbation bound (\cref{lemma:pscd_upper}).

\cref{thm:dpscd_conv} constitutes the first analysis of coordinate descent in the differentially private setting and we hope it serves as a stepping stone for future theoretical results.

\section{Experiments} \label{sec:exp}

Our empirical results compare our new \dpscd algorithm against SCD, SGD and \dpsgd.
We also include \dpsscd as a baseline, to depict the importance of independent updates within a given mini-batch.
\dpsscd adopts the natural SCD method of performing sequential  (and thus correlated) updates. 
We prove in \cref{sec:appendix_seq} that the update correlation drives the sensitivity to ${S_f \leq 2C \sqrt{L(L+1)}}$ and thus makes the noise addition significantly larger than the one for \dpscd.

We test the performance on three popular GLM applications, namely ridge regression, logistic regression and $L_2$-regularized SVMs\footnote{The hinge loss is not smooth and thus SVMs do not fall into the setup of our convergence analysis. 
Therefore, our experiments regarding SVMs illustrate that our convergence regime may not be tight and further function classes may be worth exploring.}.
We provide detailed information regarding our setup in \cref{sec:appendix_experiments}.
In particular, we describe the datasets (YearPredictionMSD, Phishing, Adult), the metrics, the values for each hyperparameter, and the deployment setup of our implementation.

\subsection{Results}

\begin{figure*}[!t]
\centering 
\subfloat[Ridge regression (YearPredictionMSD)]{\includegraphics[width=0.33\linewidth,keepaspectratio]{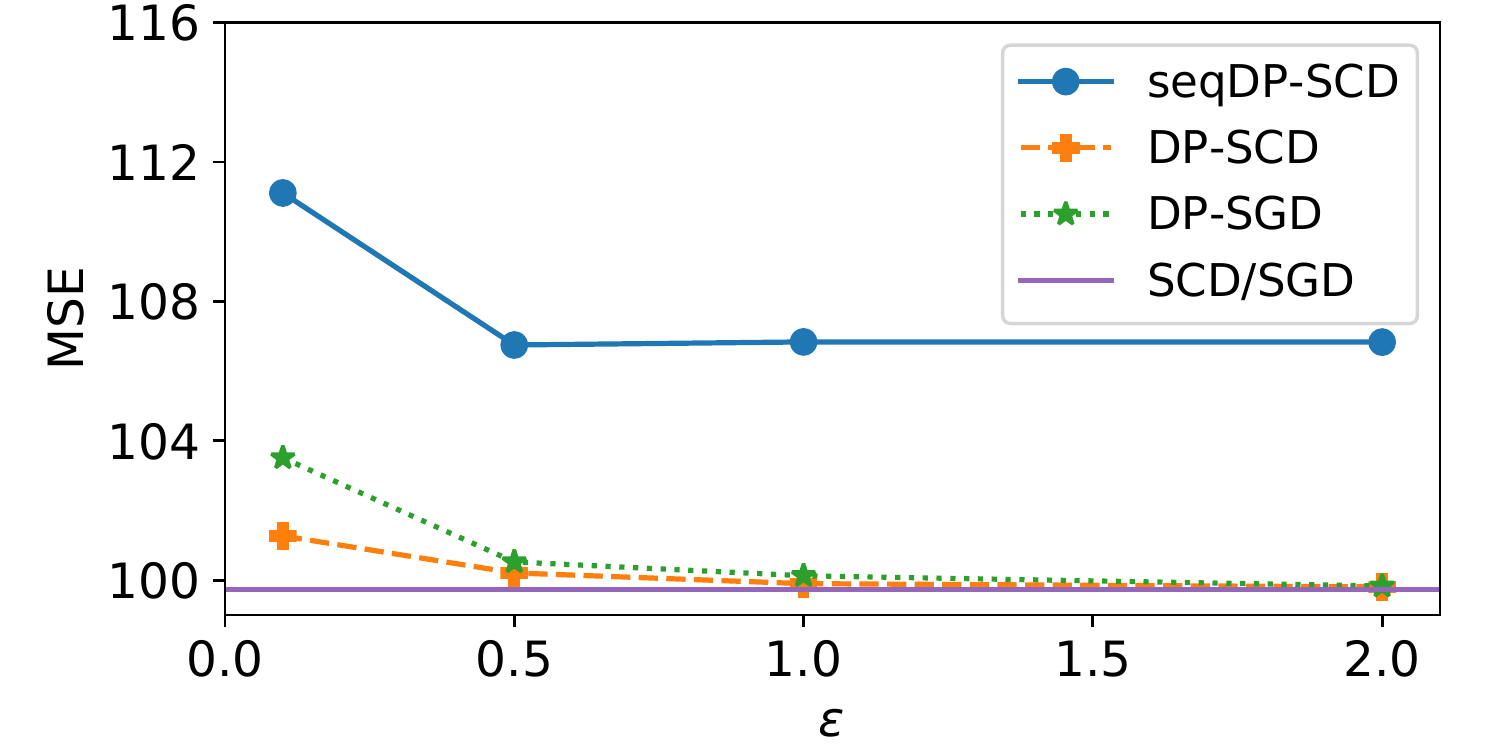} \label{fig:puMSD}} 
\subfloat[Logistic regression (Phishing)]{\includegraphics[width=0.33\linewidth,keepaspectratio]{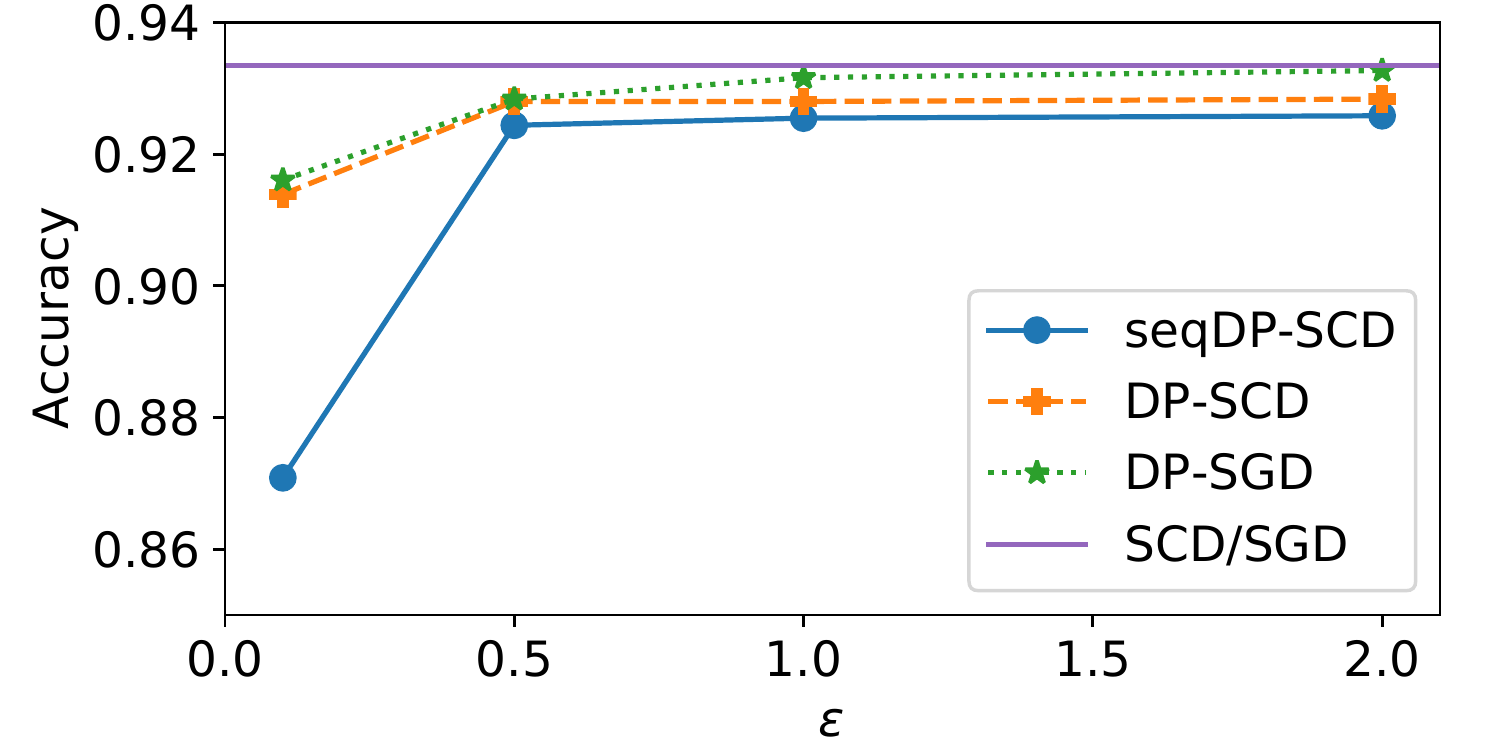} \label{fig:puPhish}}
\subfloat[SVMs (Adult)]{\includegraphics[width=0.33\linewidth,keepaspectratio]{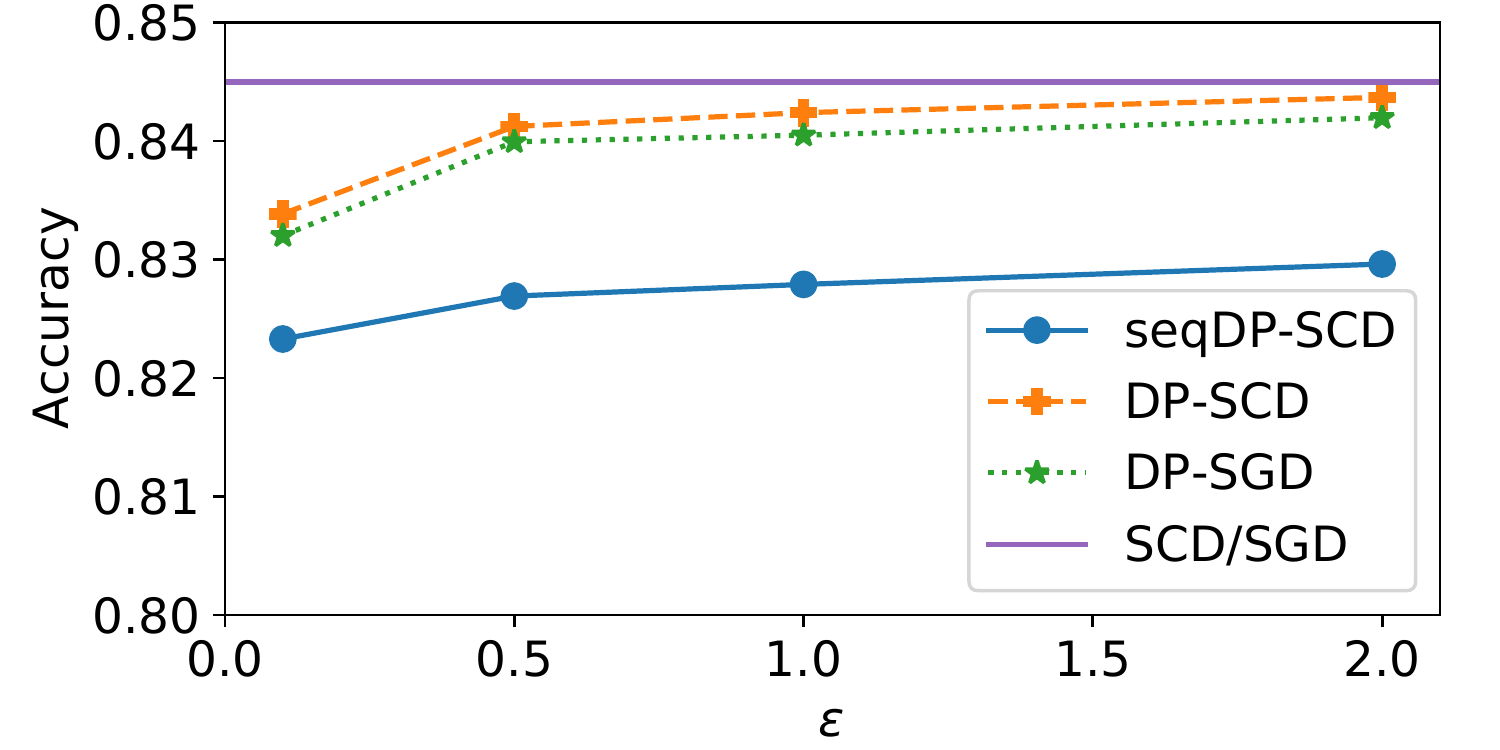} \label{fig:puSVM}}
\caption{Privacy-utility trade-off. Better utility means lower MSE or larger accuracy. \dpscd outperforms \dpsgd for the applications that enable exact update steps (namely ridge regression and SVMs). %
 }
\label{fig:pu}
\end{figure*}

\paragraph{Tuning cost.}
The hyperparameters of \dpsscd, \dpscd, SGD and \dpsgd are $\{L, C\}$, $\{L, C\}$, $\{\eta, |\xi|\}$ and $\{\eta, L, C\}$ respectively; SCD requires no tuning.
We tested a total of 156 configurations for \dpscd as opposed to a total of 2028 for \dpsgd.
For large-scale datasets that require significant amount of training time and resources, the difference in the number of hyperparameters constitutes an appealing property of \dpscd.
Noteworthy, the SGD tuning is not useful for the \dpsgd tuning, as the best choice for $\eta$ depends on the choice of $C$~\cite{thakkar2019differentially}.

\paragraph{Privacy-utility trade-off.}

\cref{fig:pu} quantifies the trade-off between privacy and utility for different privacy levels, i.e., different $\epsilon$ values under a fixed $\delta=0.001$~\cite{wang2017differentially,zhang2017efficient}.
We observe that \dpsscd has the worst performance due to the significantly larger noise.
\dpscd performs better than \dpsgd for ridge regression and SVMs, and worse for logistic regression, as the steps of \dpscd in the case of ridge regression and SVMs are more precise despite suffering more noise than \dpsgd.
Moreover, \dpscd requires $\sqrt{2}$ more noise than \dpsgd (for the same privacy guarantee) due to the need of a shared vector (\cref{sec:dpscd}).
However, each update of \dpscd finds an exact solution to the minimization problem for ridge regression and SVMs (and an approximate one for logistic regression), whereas \dpsgd takes a direction opposite to the gradient estimate.
In SGD, we need to often be conservative (e.g., choose a small learning rate) in order to account for bad estimates (due to sampling) of the actual gradient.

\begin{figure}[!t]
\centering 
\includegraphics[width=0.9\linewidth,keepaspectratio]{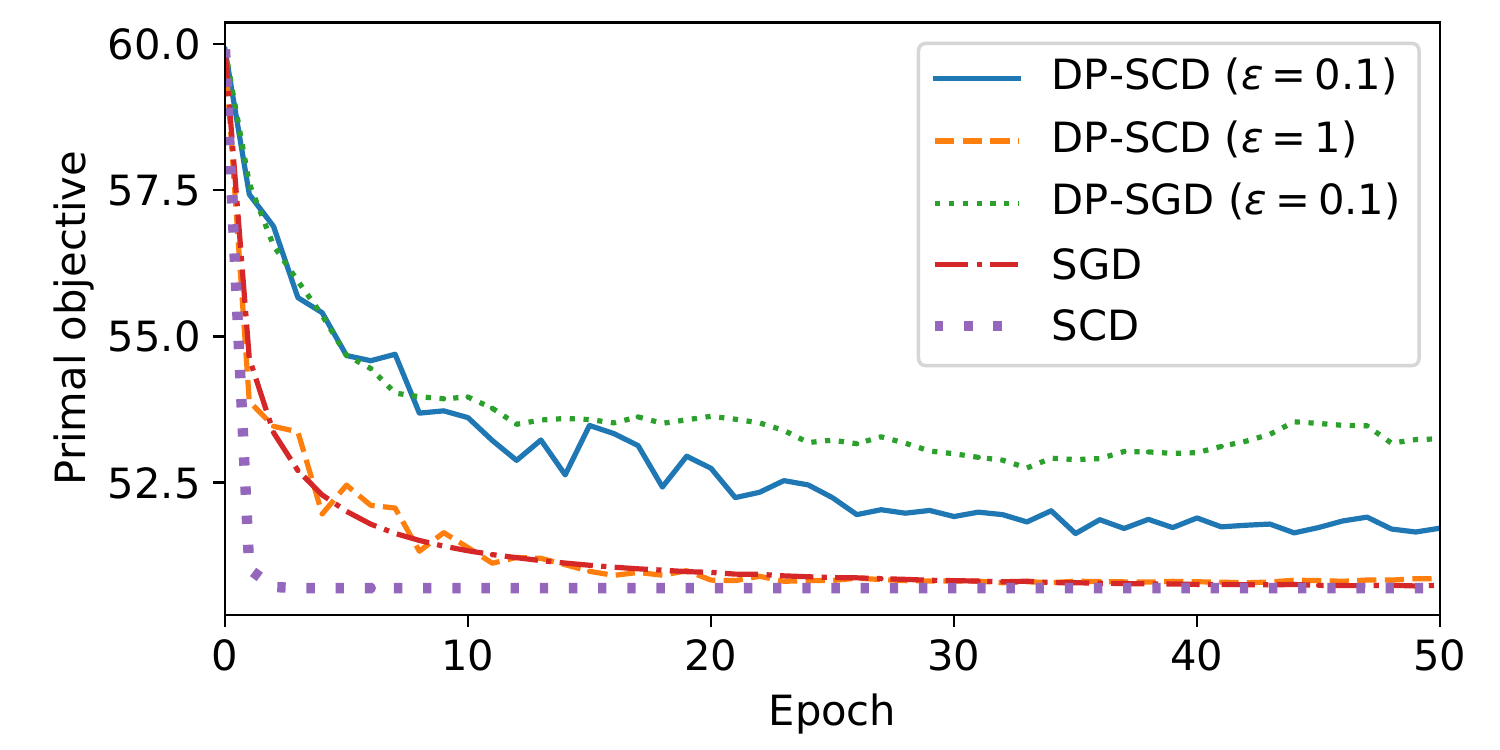} 
\caption{Impact of noise on convergence. Differential privacy does not prevent convergence but makes the objective reduction more noisy while also increases the distance to the optimum (aligned with the result of \cref{thm:dpscd_conv}).} 
\label{fig:conv}
\vspace{-4mm}
\end{figure}

\paragraph{Convergence.}
\cref{fig:conv} shows the impact of noise on the convergence behavior on the YearPredictionMSD dataset.
In particular, for a given $\epsilon$, we select the best (in terms of validation MSE) configuration (also used in \cref{fig:puMSD}), and measure the decrease in the objective with respect to epochs (not time as that would be implementation-dependent).
We empirically verify the results of \cref{thm:dpscd_conv} by observing that the distance between the convergence point and the optimum depends on the level of privacy.
Moreover, \dpscd and \dpsgd converge with similar speed for $\epsilon=0.1$, but to a different optimum (as also shown in \cref{fig:puMSD}).
Decreasing the amount of noise ($\epsilon=1$), makes \dpscd converge almost as fast as SGD and with more stability compared to $\epsilon=0.1$.
This is aligned with the results of \cref{sec:analysis}, i.e., the fact that the larger amount of noise (decrease in $\epsilon$) makes the decrease in the suboptimality more noisy.

\begin{figure}[!t]
\centering 
\includegraphics[width=0.9\linewidth,keepaspectratio]{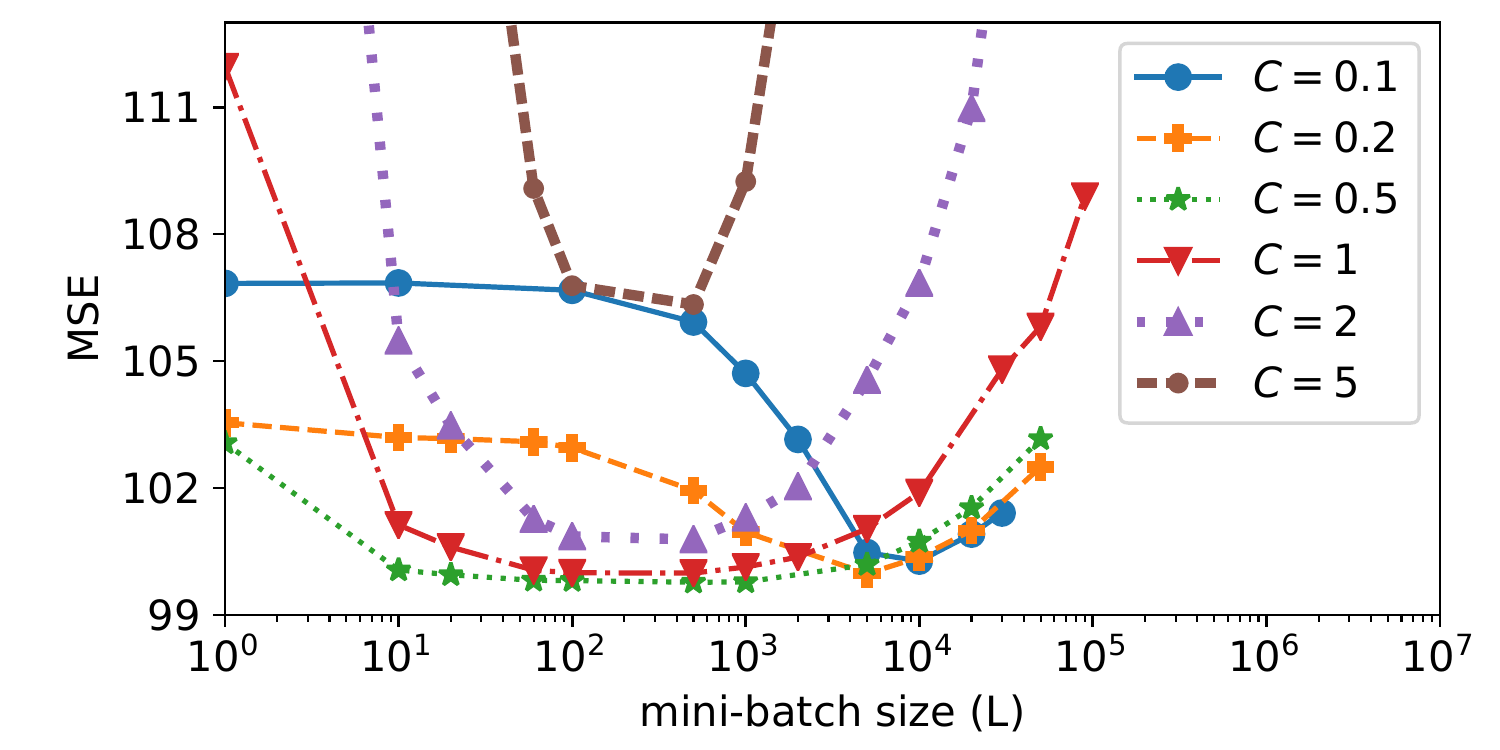} 
\caption{Impact of the mini-batch size ($L$) and the scaling factor ($C$) on the utility for \dpscd. Deviating from the best choice for $C$ ($C=0.5$ for this setup), reduces the width of the flat area and moves the minimum to the right (for smaller $C$ values) or upwards (for larger $C$ values).} 
\label{fig:lot_MSD}
\vspace{-4mm}
\end{figure}

\paragraph{Mini-batch size and scaling factor interplay.}
\dpscd involves two hyperparameters, namely the mini-batch size ($L$) and the scaling factor ($C$).
Both hyperparameters affect the privacy-utility trade-off while the mini-batch size also controls the level of parallelism (\cref{sec:dpscd}).
\cref{fig:lot_MSD} shows the impact of the choice of $L, C$ to the utility for the YearPredictionMSD dataset with $\epsilon=1$. 
For the given setup, the best (i.e., leading to the lowest MSE) value for $C$ is 0.5.
We observe that for this value the curve also flattens out, i.e., there is a wide range of values for $L$ that achieve MSE close to the lowest.
On the one hand, decreasing $C$ reduces the width of the flat area around the minimum and shifts the minimum to the right (i.e., to larger mini-batch sizes).
This can be attributed to the fact that larger $C$ corresponds to smaller updates in terms of magnitude (Step~8 in \cref{algo:dp-pscd_dual}) and thus aggregating more of them (i.e., updating with larger mini-batch size) is necessary to minimize the MSE.
On the other hand, increasing $C$ also reduces the width of the flat area while also increases the MSE.
The updates in this case have larger magnitudes (and thus require larger noise) thereby preventing \dpscd from reaching a good minimum.

\begin{table*}[t!]
\small
\centering
\begin{tabular}{|c|c|c|}
\hline
 Method & Perturbation & Utility Bound \\ \hline
 \cite{zhang2017efficient} & Output              & $\bigO\left(\frac{M}{N^2\epsilon^2}\right)$ \\ \hline
 \cite{chaudhuri2009privacy,chaudhuri2011differentially} & Inner (objective)           & $\bigO\left(\frac{M}{N^2\epsilon^2}\right)$ \\ \hline
 \cite{wang2017differentially} & Inner (update)            & $\bigO\left(\frac{M \cdot \log(N)}{N^2\epsilon^2}\right)$ \\ \hline
 \dpscd & Inner (update)            & $\bigO\left(\frac{L^3 \cdot \log(\frac{N}{L})}{N^4 \epsilon^2}\right)$ \\ \hline
\end{tabular}
\caption{Comparison of utility bounds of (${\epsilon}, {\delta}$)-DP algorithms for empirical risk minimization.}
 \label{table:dperm}
 \vspace{-4mm}
\end{table*}

\section{Related Work} \label{sec:relworks}

\paragraph{Perturbation methods.}
Existing works achieve differentially private ML 
by perturbing the query output (i.e., model prediction).
These works target both convex and non-convex optimization and focus on a specific application~\cite{chaudhuri2009privacy,nguyen2017differentially}, a subclass of optimization functions (properties of the loss function)~\cite{chaudhuri2011differentially} or a particular optimization algorithm~\cite{abadi2016deep,talwar2015nearly}.
These approaches can be divided into three main classes.
The first class involves \emph{input perturbation} approaches that add noise to the input data~\cite{duchi2013local}.
These approaches are easy to implement but often prohibit the ML model from providing accurate predictions.
The second class involves \emph{output perturbation} approaches that add noise to the model after the training procedure finishes, i.e., without modifying the vanilla training algorithm. 
This noise addition can be model-specific~\cite{wu2017bolt} or model-agnostic~\cite{bassily2018model,papernot2018scalable}.
The third class involves \emph{inner perturbation} approaches that modify the learning algorithm such that the noise is injected during learning.
One method for inner perturbation is to modify the objective of the training procedure~\cite{chaudhuri2011differentially}. 
Another approach involves adding noise to the output of each update step of the training without modifying the objective~\cite{abadi2016deep}. Our new \dpscd algorithm  belongs to the third class.

\paragraph{DP for Empirical Risk Minimization (ERM).}
Various works address the problem of ERM (similar to our setup \cref{sec:setup}), through the lens of differential privacy.
\cref{table:dperm} compares the utility bounds between \dpscd and representative works for each perturbation method for DP-ERM.
We simplify the bounds following~\cite{wang2017differentially} for easier comparison.
The assumptions of these methods, described in~\cite[Table~1]{wang2017differentially} and \cref{sec:analysis}, are similar\footnote{\dpscd does not require the loss function to be Lipschitz.}.
We highlight that the bound for \dpscd is independent of the dimensionality of the problem ($M$) due to the dual updates, while also includes the mini-batch size ($L$) for quantifying the impact of the varying degree of parallelism.
If $L=\bigO(N)$, then the ratio $N/M$ determines whether the bound of \dpscd is better (i.e., smaller).
We plan to investigate the effect of this ratio in our future work.

Existing DP-ERM methods based on SGD typically require the tuning of an additional hyperparameter (learning rate or step size) as in \dpsgd~\cite{abadi2016deep}.
The value of this hyperparameter for certain loss functions can be set based on properties of these functions~\cite{wu2017bolt}.
Furthermore regarding~\cite{wu2017bolt}, the authors build upon permutation-based SGD and employ output perturbation, but tolerate only a constant number of iterations.

\paragraph{Coordinate descent.}
SCD algorithms do not require parameter tuning if they update one coordinate (or block) at a time, by exact minimization (or Taylor approximation).
One such algorithm is SDCA~\cite{shalev2013stochastic} that is similar to \dpscd when setting $\epsilon\rightarrow \infty,L=1$ and $C\rightarrow \infty$.
Alternative SCD algorithms take a gradient step in the coordinate direction that requires a step size~\cite{nesterov2012efficiency,beck2013convergence}.

Parallelizable variants of SCD such as~\cite{bradley2011parallel,richtarik2016parallel,syscd} have shown remarkable speedup when deployed on multiple CPUs/GPUs~\cite{parnell2017large,hsieh2015passcode,shalev2013accelerated,chiang2016parallel,zhuang2018naive}, or multiple machines~\cite{ma2015adding,dunner2018snap}.
These works employ sampling to select the data to be updated in parallel.
\dpscd also employs sampling via the mini-batch size ($L$), similar to the lot size of \dpsgd~\cite{abadi2016deep}. This enables parallel updates and improves the privacy-utility trade-off.
\dpscd also builds on ideas from distributed learning, such as the CoCoA method~\cite{ma2015adding} where in our case each node computes its update based on a single datapoint from its data partition.

\section{Conclusion} \label{sec:conclusion}

This paper presents the first differentially private version of the popular stochastic coordinate descent algorithm.
We demonstrate that extending SCD with a mini-batch approach is crucial for the algorithm to be competitive against SGD-based alternatives in terms of privacy-utility trade-off.
To achieve reliable convergence for our mini-batch parallel SCD, we build on a separable surrogate model to parallelize updates across coordinates, inspired by block separable distributed methods such as in~\cite{smith2017cocoa}. 
This parallel approach inherits the strong convergence guarantees of the respective method.
In addition we provide a utility guarantee for our DP-SCD algorithm despite the noise addition and the update scaling.
We also argue that the dual formulation of \dpscd is preferable over the primal due to the example-wise access pattern of the training data, that is more aligned with the focus of differential privacy (i.e., to protect individual entries in the training data).
Finally, we provide promising empirical results for DP-SCD (compared to the SGD-based alternative) for three popular applications.

\clearpage

\section*{Acknowledgements}

CM would like to acknowledge the support from the Swiss National Science Foundation (SNSF) Early Postdoc.Mobility Fellowship Program.

\section*{Ethical Impact}

Many existing industrial applications are based on the SCD algorithm while they employ sensitive data (e.g., health records) to train machine learning models. 
Our algorithm (DP-SCD) provides a direct mechanism to improve the privacy-utility trade-off while also preserves the core structure of SCD.
Therefore, big industrial players can extend existing implementations (that are the outcome of a large amount of research and engineering efforts) to safeguard the privacy of the users up to a level that preserves the utility requirements of the application.

\bibliography{biblio}

\onecolumn

\begin{center}
{\LARGE \bf Differentially Private Stochastic Coordinate Descent \\ (technical appendix)} 
\end{center}

\begin{appendices}
\section{\dpscd}

\subsection{Cost Analysis} \label[app]{sec:appendix_cost}

The performance overhead of \dpscd with respect to SCD boils down to the cost of sampling the Gaussian distribution.
This cost is proportional to the mini-batch size $L$ -- larger $L$ means less frequent noise additions and thus less frequent sampling.
The time complexity for \cref{algo:dp-pscd_dual} is $\bigO(T   M)$. 
The updates for the coordinates within a given mini-batch can be parallelized; we discuss parallelizable variants of SCD in \cref{sec:relworks}.
Note that after the noise addition, even for sparse data, the resulting updates to $\vv$ are no longer sparse. 
This prohibits any performance optimizations that rely on data sparsity for accelerating training.

\subsection{Implementation Details} \label[app]{sec:appendix_implementation}

In our implementation we use the moments accountant to determine a tight bound on the privacy budget according to Step~2 of \cref{algo:dp-pscd_dual}. 
In particular, we choose the smallest $\sigma$ that provides the given privacy guarantee, i.e.,  
$\text{MA}\left(\delta, \sigma, \frac L N , T\right) \leq \epsilon$, 
where $T$ denotes the number of iterations of \cref{algo:dp-pscd_dual}.
Given that $\epsilon$ decreases monotonically with increasing $\sigma$, we find $\sigma$ by performing binary search until the variance of the output of the MA gets smaller than 1\% of the given $\epsilon$.

The update computation (Step~7 in \cref{algo:dp-pscd_dual}) can involve dataset-dependent constraints for applications such as logistic regression or SVMs.
For example, logistic regression employs the labels to ensure that the logarithms in the computation of $\zeta_j$ are properly defined~\cite{shalev2013stochastic}.
An approach that enforces these constraints after the noise addition would break the privacy guarantees as it employs the sensitive data.
We thus enforce these constraints before computing $\zeta_j$. 
As a result, the output model does not respect these constraints but there are no negative implications as the purpose of these constraints is to enable valid update computations.

\subsection{Proof of \cref{lemma:sens-dppscd}} \label[app]{sec:appendix_dppscd_sens}

\sensdppscd* 

\begin{proof} The function $\fv$ takes the model vector $\alphav$ and the auxiliary vector $\vv$ as input, and outputs the updated vectors $\alphav+\Delta \alphav$ and $\vv+\Delta\vv$.
To analyze the sensitivity of $\fv$ we consider two data matrices $\Xv, \Xv'$  that differ only in a single example $\xv_k$. The information of $\xv_k$ is removed from $\Xv'$ by initializing $\Xv'=\Xv$ and setting its $k$-th column to zero. Let us denote the updates computed on $\Xv$ as $(\Delta \vv, \Delta\alphav)$ and the updates computed on $\Xv'$ as $(\Delta \vv', \Delta\alphav')$.
First, to analyze the updates to $\vv$ we note that the scaling operation in Step~8 of \cref{algo:dp-pscd_dual} guarantees $|\zeta_k| \leq C$ and by normalization we have ${\|\xv_k \| = 1}$  which yields $\|\Delta\vv-\Delta\vv' \| \leq  \|\xv_k\|  |\zeta_k| \leq C$. Similarly, given that the updates to $\alphav$ for each coordinate in the mini-batch $\cB$ are computed independently,  we have  $\|\Delta\alphav-\Delta\alphav'\|=|\zeta_k| \leq C$. 
Hence, the overall sensitivity of $\fv$ operating on the concatenated vector $[\alphav; \vv]$  is bounded by
 $S_f^2 := \max\limits_{\Xv \setminus \Xv' = \xv_k}  \|\Delta\vv-\Delta\vv' \|^2 + \|\Delta\alphav-\Delta\alphav'\|^2\leq  2C^2$.

\end{proof}

\section{Primal Version} \label[app]{sec:appendix_primal}

\begin{algorithm}[H]
\caption{\pdpscd (for \cref{eq:obj})} \label{algo:dp-pscd_primal}
\small
\DontPrintSemicolon
\KwIn{$S$: sample size, same input as \cref{algo:dp-pscd_dual}}

\textbf{Init:} $\thetav= \bm{0}$ ; $\vv = \bm{0}$ ; shuffle examples $\xv_i$ \;

$\sigma \leftarrow \textit{smallest noise magnitude, s.t., } \text{MA}\left(\delta, \sigma, 1, T\right) = \epsilon$ \; 

\For{$t = 1, 2, \cdots T $}{
  $\Delta\vv = \bm{0}$ ; $\Delta\thetav = \bm{0}$\;
  Randomly sample a mini-batch $\mathcal{B} \subset [M]$ of $L$ coordinates  \;
  \For{$j \in \cB$}{

  $\zeta_j = \argmin_\zeta \cG_j(\theta_j,\zeta, \vv, \Xv[j,:])$ \;  
  $\zeta_j \mathbin{{/}{=}} \max\left(1, \frac{|\zeta_j|}{C}\right)$ \tcp*[r]{scale} 

  ${\Delta \vv} \; \mathbin{{+}{=}} \zeta_j \Xv[j,:] $ \;

  $\Delta\thetav\; \mathbin{{+}{=}} \ev_j \zeta_j$  
  }

  \tcp*[l]{update the model} 
  $\thetav\; \mathbin{{+}{=}} \Delta\thetav + \ev_{\cB} ( \cN(0, \sigma^2 4C^2L(L+1) \bm{I}_L))$, \quad $\bm{I}_L \in \bbR^{L \times L}$ \;
  $\vv\; \mathbin{{+}{=}} \Delta\vv + \cN(0, \sigma^2 4C^2L(L+1) \bm{I}_N)$, \quad $\bm{I}_N \in \bbR^{N \times N}$ \;

}

\Return $\thetav$ \tcp*[r]{primal DP model}
\end{algorithm}

\begin{restatable}[Sensitivity of \pdpscd]{lem}{sensdppscdprimal} \label{lemma:sens-dppscd_primal} Assume the rows of the data matrix $\Xv$ are normalized.
Then, the sensitivity of each mini-batch update computation (Steps~6-11 in \cref{algo:dp-pscd_primal}) is bounded by: ${S_f \leq 2C \sqrt{L(L+1)}}$.
\end{restatable}

\begin{proof} The function $\fv$ takes the model vector $\thetav$ and the auxiliary vector $\vv$ as input, and outputs the updated vectors $\thetav+\Delta \thetav$ and $\vv+\Delta\vv$ (Steps 6-11 in \cref{algo:dp-pscd_primal}).
Let us consider the sensitivity of $\fv$ with respect to a change in a single example $\xv_k$, therefore let $\Xv'$ be defined as the matrix $\Xv$ where the example $\xv_k$ is removed by setting the $k$-th column of $\Xv'$ to zero.  
Let $(\Delta \vv, \Delta\thetav)$ denote the updates computed on $\Xv$ and $(\Delta \vv', \Delta\thetav')$ the updates computed on $\Xv'$.
A crucial difference for computing the sensitivity, compared to the dual version, is that the missing data vector $\xv_k$ can affect all $L$ coordinate updates to the model vector~$\thetav$. 
In the worst case the missing data point alters the sign of each coordinate update $j\in\cB$ such that $[\Delta \thetav- \Delta \thetav']_{j}=2 |\zeta_{j}|=2C$. The bound on $|\zeta_{j}|$ is given by the scaling Step~8 of \cref{algo:dp-pscd_primal}. 
This yields $\|\Delta \thetav- \Delta \thetav'\|^2\leq 4 L C^2$.
Focusing on the auxiliary vector, we note that  $\Delta\vv-\Delta\vv' = \sum_{j\in\cB} \Delta \theta_j \Xv_{j,:} - \Delta \theta'_j  \Xv'_{j,:}$.  
Assuming the rows of $\Xv$ are normalized (i.e., $\|X_{j,:}\|=1\;\forall j$) we get  $\|\Delta\vv-\Delta\vv'\| \leq \sum_{j\in\cB} |\Delta \theta_j |\| \Xv_{j,:}\| +|\Delta \theta'_j |\| \Xv'_{j,:}\| \leq 2 L C$.  
Hence, the sensitivity for the concatenated output vector $[\thetav; \vv]$ of the mini-batch update computation $\fv$ is bounded as
$S_f^2 := \max\limits_{\Xv \setminus \Xv' = \xv_k} \| \fv(\Xv) - \fv(\Xv') \|^2=  \max\limits_{\Xv \setminus \Xv' = \xv_k} \|\Delta \vv- \Delta \vv'\|^2+\|\Delta \thetav- \Delta \thetav'\|^2\leq   4 LC^2(L+1)$.

\end{proof}

\section{Sequential Version} \label[app]{sec:appendix_seq}

In this section we present a baseline algorithm that we call \dpsscd to depict the importance of adopting the subproblem formulation of~\cite{ma2015adding} to create independent updates inside a given mini-batch, for \dpscd.
\dpsscd, outlined  in \cref{algo:dp-sscd_dual}, adopts the natural SCD method of performing \emph{sequential} and thus correlated updates within a given mini-batch.
In particular, the updates for both $\alphav$ and $\vv$ at sample $j$ (Steps~8-9 of \cref{algo:dp-sscd_dual}) depend on all the previous coordinate updates within the same mini-batch ($\cB$).
In contrast, \dpscd (\cref{algo:dp-pscd_dual}) eliminates these correlations by computing all updates independently.

On the one hand, correlated, sequential updates are better for convergence in terms of sample complexity~\cite{hsieh2015passcode}, but on the other hand these correlations require significantly more noise addition than \dpscd (\cref{algo:dp-pscd_dual}) to achieve the same privacy guarantees (see \cref{lemma:sens-dpsscd}).
The difference in the amount of noise makes the overall performance of \dpscd superior to \dpsscd as we empirically confirm in \cref{sec:exp}.

\begin{algorithm}[h!]
\caption{\dpsscd (for \cref{eq:objD})} \label{algo:dp-sscd_dual}
\small
\DontPrintSemicolon
\tcp*[l]{same as Steps~1-2 of \cref{algo:dp-pscd_dual}}
\setcounter{AlgoLine}{2} 
\For{$t = 1,2,...T $}{
  Randomly sample a mini-batch $\mathcal{B} \subset [N]$ of $L$ examples \;
  \For{$j \in \cB$}{

    $\zeta_j = \argmin_\zeta \cG_j^*(\alpha_j,\zeta, \vv, \xv_j)$ \; 
    $\zeta_j \mathbin{{/}{=}} \max\left(1, \frac{|\zeta_j|}{C}\right)$ \tcp*[r]{scale}

    $\alphav \; \mathbin{{+}{=}} \ev_j \zeta_j$  \tcp*[r]{update the model}
    $\vv \; \mathbin{{+}{=}} \zeta_j \xv_j $ \;
   }

  \tcp*[l]{add noise} 
  $\alphav\; \mathbin{{+}{=}} \ev_\cB \cN(0, \sigma^2 4C^2L(L+1) \bm{I_L})$, \quad $\bm{I_L} \in \bbR^{L \times L}$ \;
  $\vv\; \mathbin{{+}{=}} \cN(0, \sigma^2 4C^2L(L+1) \bm{I_M})$, \quad $\bm{I_M} \in \bbR^{M \times M}$\;
}
\Return $\bm{\theta} =\frac 1 {\lambda N}\vv$ \tcp*[r]{primal DP model}
\vspace{0.1mm}
\end{algorithm}

\begin{restatable}[Sensitivity of \dpsscd]{lem}{sensdpsscd} \label{lemma:sens-dpsscd}
Assume the columns of the data matrix $\Xv$ are normalized. Then, the sensitivity of each update computation Step~5-10 in \cref{algo:dp-sscd_dual} is bounded by 
\[{S_f \leq 2C \sqrt{L(L+1)}} \]
\end{restatable}
\begin{proof}
The proof is similar to \cref{lemma:sens-dppscd_primal}.
The difference among $\fv(\Xv), \fv(\Xv')$ consists of (a) the difference due to the missing example for $\Xv'$ and (b) the difference due to all the subsequent values (correlated updates).
Moreover, the subsequent values-vectors can, in the worst case, be opposite. 
Therefore, the sensitivity follows by using the triangle inequality.
\end{proof}

\section{Convergence Analysis}

\subsection{Proof of \cref{lemma:pscd_lower}} \label[app]{sec:proof_pscd_lower}

\pscdlower*

\begin{proof}

Consider the update step (a) that employs a set of unscaled updates $\Delta\alphav^\text{tmp}$. The coordinate updates $\Delta\alphav_j^\text{tmp}=\zeta_j$ are computed by minimizing $\cG_j^*$ for all $ j\in\cB$ in Step~7 of \cref{algo:dp-pscd_dual}. Given the definition of $\cG^*_j$ in \cref{eq:Fhat}, the decrease in the dual objective is lower-bounded as:
\begin{align*}
\Delta_F:=  \cF^*(\alphav) - \cF^*(\alphav+\Delta\alphav^{\text{tmp}})  
\geq \cF^*(\alphav) - \sum_{i=1}^L \cG_j^*(\alpha_j,\zeta_j, \vv, \xv_j) \;\;\forall \zeta_j \in \bbR
\end{align*}
which follows from \cite[Lemma~3]{ma2015adding} with parameters $\gamma=1$ and $\sigma'=L$. Plugging in the definition of $\cF^*$ (\cref{eq:objD}) and $\cG^*_j$ (\cref{eq:Fhat}) this yields:
\begin{equation}
\Delta_F \geq \frac 1 N \sum_{j=1}^L \ell_j^*(-\alpha_j) + \frac 1 {2\lambda N^2}\|\vv  \|^2 - \frac 1 N \sum_{j=1}^L \ell_j^*(-(\alpha_j +\zeta_j)) -  \frac 1 {2\lambda N^2} \left( \|\vv\|^2 + 2  \xv_j^\top \vv \zeta_j+  L  \|\xv_j \|^2 \zeta_j^2 \right)
\end{equation}
Note that the above bound holds for every $ \zeta_j \in \bbR$. In the following we restrict our consideration to updates of the form  $\zeta_j:= s(u_j-\alpha_j)$ for any $s \in (0,1]$ and for ${u_j = -\nabla_j\ell_j\left(\frac 1 {\lambda N} \xv_j^\top \Exp[\vv]\right)}$. This choice is motivated by the analysis in~\cite{shalev2013stochastic}  and in some sense, $s$ denotes the deviation from the ``optimal'' update-value ($u_j - a_j$).
Hence we have:
\begin{small}
\begin{eqnarray*}
 \Delta_F &\geq& \frac 1 N \left[\sum_{j=1}^L \left( \ell_j^*(-\alpha_j)- \ell_j^*(-(\alpha_j +s(u_j-\alpha_j))) \right) - \frac{L}{2\lambda N} \sum_{j=1}^L  \| \xv_j s(u_j - a_j) \|^2 - \frac{1}{\lambda N}\sum_{j=1}^L \xv_j^\top \vv s(u_j - a_j) \right]\\
&\geq& 
\frac 1 N \left[\sum_{j=1}^L \left( - s \ell_j^*(-u_j) +s \ell_j^*(-\alpha_j) + \frac \mu 2 s (1-s) (u_j-\alpha_j)^2 \right. - \left. \frac {L s^2(u_j-\alpha_j)^2} {2\lambda N} \|\xv_j\|^2  -\frac 1 {\lambda N} \xv_j^\top \vv s(u_j-\alpha_j) \right)\right] 
\end{eqnarray*}
\end{small}
where we used $\mu$-strong convexity of $\ell_i^*$ (follows from $1/\mu$-smoothness of $\ell_i$) in the second inequality.

\begin{eqnarray*}
  \Exp[\Delta_F]&\geq& \frac 1 N \sum_{j=1}^L \left( - s \Exp[\ell_j^*(-u_j)] + s \Exp[\ell_j^*(-\alpha_j)] + \frac \mu 2 s (1-s) \Exp[(u_j-\alpha_j)^2] \right.\\
&&  \left. -\frac {L s^2} {2\lambda N}\Exp[(u_j-\alpha_j)^2]  \|\xv_j\|^2  -\frac {s} {\lambda N} \xv_j^\top \Exp[\vv] (u_j-\Exp[\alpha_j]) \right) \\
&\geq& \frac 1 N \sum_{j=1}^L \left( - s \ell_j^*(-u_j) + s \Exp[\ell_j^*(-\alpha_j)] +\frac {s} {\lambda N} \xv_j^\top  \Exp[\vv] \Exp[ \alpha_j] \right. \\
&& \left. -\frac {s} {\lambda N} \xv_j^\top \Exp[\vv] u_j \right.  \left. + \left(\frac \mu 2 s (1-s)-\frac {L s^2} {2\lambda N} \|\xv_j\|^2 \right) (u_j^2-2 u_j \Exp[\alpha_j] +\Exp[\alpha_j]^2+\sigma^2) \right)  \\
\end{eqnarray*}
By observing that the Fenchel-Young inequality holds as equality given $u_j = -\nabla_j\ell_j\left(\frac 1 {\lambda N} \xv_j^\top \Exp[\vv]\right)$:
\begin{eqnarray*}
  \ell_j\left(\frac 1 {\lambda N} \xv_j^\top \Exp[\vv]\right) + \ell_j^*(-u_j)  = -\frac 1 {\lambda N} \xv_j^\top \Exp[\vv]  u_j
\end{eqnarray*}
we have
\begin{myalign} \label[ineq]{ineq:pscd_exp}
 \Exp[\Delta_F] \geq& \frac 1 N \sum_{j=1}^L \left( s \ell_j\left(\frac 1 {\lambda N} \xv_j^\top \Exp[\vv]\right)+ s \Exp[\ell_j^*(-\alpha_j)]  +\frac {s} {\lambda N} \xv_j^\top  \Exp[\vv] \Exp[ \alpha_j] \notag \right. \\
&+ \left. \left(\frac \mu 2 s (1-s)-\frac {L s^2} {2\lambda N} \|\xv_j\|^2 \right) (u_j^2-2 u_j \Exp[\alpha_j] +\Exp[\alpha_j]^2+\sigma^2) \right) 
\end{myalign}
We then employ the definition of the duality gap ${\text{Gap}(\alphav) :=  \cF(\thetav(\alphav);\Xv) - (-\cF^*(\alphav;\Xv))}$ and take the expectation w.r.t. the randomization in the noise along with the Jensen inequality for convex functions.
\begin{myalign*}
\Exp[\text{Gap}(\alphav)] &=  \cF(\thetav(\Exp[\alphav]);\Xv) + \Exp[\cF^*(\alphav;\Xv)] \\
&= \frac 1 N \sum_{j=1}^N \left(\ell_j(\xv_j^\top \thetav)+\Exp[\ell_j^*(-\alpha_j)]\right) + \frac \lambda {2} \|\thetav(\Exp[\alphav])\|^2 + \frac 1 {2\lambda N^2}\|\Xv \Exp[\alphav]\|^2 
\end{myalign*}
We then apply the Fenchel-Young inequality:
\begin{eqnarray*}
g(\thetav(\Exp[\alphav])) + g^*(\Xv \Exp[(\alphav]) &\geq& \thetav(\Exp[\alphav])^\top \Xv \Exp[\alphav] \\
\Leftrightarrow \frac \lambda 2 \|\thetav(\Exp[\alphav]) \|^2 + \frac 1 {2\lambda N^2} \|\Xv \Exp[\alphav]\|^2 &\geq& \frac 1 {\lambda N} \Exp[\vv]^\top \Xv \Exp[\alphav]
\end{eqnarray*}

The above inequality holds as equality in light of \cref{remark:pscd_consistency} and the primal-dual map.
Therefore by using uniform mini-batch sampling the duality gap becomes:
\begin{eqnarray*}
\Exp[\text{Gap}(\alphav)] &=& \frac 1 N \sum_{j=1}^N \left(\ell_j(\xv_j^\top \thetav)+\Exp[\ell_j^*(-\alpha_j)] + \frac 1 {\lambda N}\Exp[\alpha_j] \xv_j^\top \Exp[\vv]\right) \\
&=& \frac 1 L \sum_{j\in\cB} \left(\ell_j(\xv_j^\top \thetav)+\Exp[\ell_j^*(-\alpha_j)] + \frac 1 {\lambda N}\Exp[\alpha_j] \xv_j^\top \Exp[\vv]\right) 
\end{eqnarray*}
By extracting these terms in \cref{ineq:pscd_exp}, the bound simplifies:
\begin{equation}
 N  \Exp[\Delta_F] \geq
 s L  \Exp[\text{Gap}(\alphav)]+ \sum_{j\in\cB} \left(\frac \mu 2 s (1-s) \right. -\left. \frac {L s^2} {2\lambda N} \|\xv_j\|^2 \right) (u_j^2-2 u_j \Exp[\alpha_j] +\Exp[\alpha_j]^2+\sigma^2)
\end{equation}
Then by using $\|\xv_j\|^2=1$ and $\text{Gap}(\alphav)\geq \cS(\alphav)$ we get:
\begin{eqnarray*}
\Exp[\Delta_F] &=& \Exp[\cS(\alphav)]-\Exp[\cS(\alphav+ \Delta\alphav^{\text{tmp}})] \notag \\
&\geq& \frac{sL}{N} \Exp[\cS(\alphav)] + \frac {L s^2}{2\lambda N^2} \left(\frac{\mu \lambda (1-s) N}{sL} -  1 \right) \sum_{j=1}^L \left((u_j-\Exp[\alpha_j])^2+\sigma^2\right)\notag\\ 
&\stackrel{s=\frac{\mu\lambda N}{\mu\lambda N + L}}{=}& 
\frac {\mu\lambda L} {\mu\lambda N + L} \Exp[\cS(\alphav)]
\end{eqnarray*}
Thus as long as $\Exp[\alphav]$ is not equal to $\alphav^\star$ we can expect a decrease in the objective from computing an update $ \Delta\alphav^{\text{tmp}}$ based on the noisy $\alphav, \vv$.

\end{proof}

\subsection{Proof of \cref{thm:dpscd_conv}} \label[app]{sec:proof_pdscd_conv}

\pscdconv*

\begin{proof}
We reorder terms in \cref{eq:pscdupper} and subtract $\cS(\alphav)$ on both sides.
We then combine \cref{eq:pscdscaling,eq:pscdlower} and get that the suboptimality decreases per round by:

\begin{equation*} \label[ineq]{eq:perRoundDecrease}
\Exp[\cS(\alphav)-\cS(\alphav+ \Delta\alphav + \etav)] \geq \frac {(1-\kappa)\mu\lambda L} {\mu\lambda N + L}\; \Exp[\cS(\alphav)] -  \frac{L \sigma^2}{2 \lambda N^2}
\end{equation*}
At iteration $t$ we thus have:
\begin{eqnarray*}
\Exp[\cS(\alphav^{(t-1)})] - \Exp[\cS(\alphav^{(t)})] &\geq& \frac {(1-\kappa)\mu\lambda L} {\mu\lambda N + L}\; \Exp[\cS(\alphav^{(t-1)})] -  \frac {L \sigma^2} {2 \lambda N^2} \\
\Leftrightarrow \Exp[\cS(\alphav^{(t)})] &\leq& \underbrace{\left(1 - \frac {(1-\kappa)\mu\lambda L} {\mu\lambda N + L}\right)}_A \Exp[\cS(\alphav^{(t-1)})]  + \frac {L \sigma^2} {2 \lambda N^2} 
\end{eqnarray*}
We apply the previous inequality recursively and get:
\begin{eqnarray*}
\Exp[\cS(\alphav^{(T)})] &\leq& A^T \Exp[\cS(\alphav^{(0)})] + \bigO\left(\frac{L \sigma^2}{\lambda N^2}\right) \\
&\stackrel{\text{\cref{thm:sigmaBound}}}{\leq}& A^T \Exp[\cS(\alphav^{(0)})] + \bigO\left(\frac{L^3 T \;\log(1/\delta)}{\lambda N^4 \epsilon^2}\right)
\end{eqnarray*}
If we choose $L$ and $C$ such that $A  < \frac 1 2 \Leftrightarrow L (2(1-\kappa)\mu\lambda - 1) > \mu\lambda N$ and $T$ such that $T = \bigO\left(\log\left(\frac{\lambda N^4 \epsilon^2}{L^3 \; \log(1/\delta)}\right)\right)$, we get the bound on the utility:
\begin{equation*}
\Exp[\cS(\alphav^{(T)})] \leq \bigO\left(\frac{L^3 \; \log(1/\delta)}{\lambda N^4\epsilon^2}\right) + \bigO\left(\frac{L^3 T \;\log(1/\delta)}{\lambda N^4 \epsilon^2}\right) 
\end{equation*}
By omitting the $\log$ term the bound on T simplifies as: ${T = \bigO\left(\log\left(\frac{\lambda N \epsilon}{L}\right)\right)}$. Hence the utility bound becomes:
\begin{equation*}
\Exp[\cS(\alphav^{(T)})] \leq \bigO\left(\frac {L^3} {\lambda N^4 \epsilon^2} \; \log\left(\frac{\lambda N\epsilon}{L}\right) \; \log\left(\frac 1 \delta\right) \right)
\end{equation*}

\end{proof}

\section{Experimental Setup} \label[app]{sec:appendix_experiments}

\paragraph{Datasets.}
We employ public real datasets.
In particular, we report on YearPredictionMSD\footnote{\url{https://www.csie.ntu.edu.tw/~cjlin/libsvmtools/datasets/regression.html}} for ridge regression, Phishing\footnote{\url{https://www.csie.ntu.edu.tw/~cjlin/libsvmtools/datasets/binary.html}} for logistic regression, and Adult\footnote{\url{https://archive.ics.uci.edu/ml/datasets/Adult}} for SVMs.
We preprocess each dataset by scaling each coordinate by its maximum absolute value, followed by scaling each example to unit norm (normalized data).
For YearPredictionMSD we center the labels at the origin.
Based on~\cite{berk2017convex} and regarding Adult, we convert the categorical variables to
dummy/indicator ones and replace the missing values with the most frequently occurring value of the corresponding feature.
We employ a training/test split for our data, to train and test the performance of our algorithms.
YearPredictionMSD and Adult include a separate training and test set file.
Phishing consists of a single file that we split with 75\%:25\% ratio into a training and a test set.
Finally, we hold-out a random 25\% of the training set for tuning the hyperparameters (validation set).
The resulting training/validation/test size is \{347786/115929/51630, 24420/8141/16281, 6218/2073/2764\} and the number of coordinates are \{90, 81, 68\} for \{YearPredictionMSD, Adult, Phishing\} respectively.

\paragraph{Performance metrics.}
\emph{Accuracy} measures the classification performance as the fraction of correct predictions among all the predictions.
The larger the accuracy, the better the utility.
\emph{Mean squared error (MSE)} measures the prediction error as: 
${\text {MSE} = \frac{1}{N} \sum_{i=1}^{N}(\hat{y_i} - y_i)^2}$
where $\hat{y_i}$ is the predicted value and $y_i$ is the actual value.
The lower the MSE, the better the utility.
We quantify convergence by showing the decrease in the primal objective ($\cF(\thetav; \Xv^{\text{(training)}})$ from \cref{eq:obj}) on the training set.

\paragraph{Hyperparameters.}
We fix $\lambda$ to $10^{-4}$ for YearPredictionMSD and Phishing and to $10^{-5}$ for the Adult dataset based on the best performance of SCD and SGD for a range of ${\lambda \in \{10^{-8}, 10^{-7}, \cdots, 1, 10, \cdots, 10^8\}}$.
For a fair comparison of the DP algorithms, the iterations need to be fixed.
Based on~\cite{wu2017bolt}, we test the DP algorithms for $\{5, 10, 50\}$ epochs and fix the number of iterations to $T=50N$ (i.e., 50 epochs) for YearPredictionMSD and $T=10N$ for the other datasets.
Based on~\cite{wang2017differentially,zhang2017efficient}, we vary $\epsilon$ in $\{0.1, 0.5, 1, 2\}$ and fix $\delta=0.001$.
We choose the other hyperparameters by selecting the combination with the best performance (lowest MSE for ridge regression and largest accuracy for logistic regression and SVMs) on the validation set.
The range of tested values is as follows.
\begin{itemize}
\item $C, \eta \in \{10^{-8}, 10^{-7}, \cdots, 1, \cdots, 10^4 \}$
\item $|\xi|, L \in \{0, 5, 10, 50, 100, 200, 500, 1000, 1250, 1500, 1750, 2000\}$
\end{itemize}

\paragraph{Deployment.}
We run our experiments on a commodity Linux machine with Ubuntu 18.04, an Intel Core i7-1065G7 processor and 32 GB of RAM.
There are no special hardware requirements for our code other than enough RAM to load the datasets.
The software versions along with the instructions for running our code are available in the README file in our code appendix.
We report the median result across 10 different runs by changing the seeding, i.e., the randomization due to initialization, sampling and Gaussian noise.

\end{appendices}

\end{document}